\documentclass{article}
\pdfoutput=1

\PassOptionsToPackage{numbers, compress}{natbib}

\usepackage[final]{neurips_bdl2019}

\usepackage[utf8]{inputenc} % allow utf-8 input
\usepackage[T1]{fontenc}    % use 8-bit T1 fonts
\usepackage{hyperref}       % hyperlinks
\usepackage{url}            % simple URL typesetting
\usepackage{booktabs}       % professional-quality tables
\usepackage{amsfonts}       % blackboard math symbols
\usepackage{nicefrac}       % compact symbols for 1/2, etc.
\usepackage{microtype}      % microtypography

\usepackage{tikz}
\usepackage{pgfplots}

\usetikzlibrary{calc}
\usetikzlibrary{positioning}
\usepgflibrary{fpu}

\usetikzlibrary{external}
\pgfplotsset{compat=newest}

\newenvironment{customlegend}[1][]{%
    \begingroup
    \csname pgfplots@init@cleared@structures\endcsname
    \pgfplotsset{#1}%
}{%
    \csname pgfplots@createlegend\endcsname
    \endgroup
}%
\def\addlegendimage{\csname pgfplots@addlegendimage\endcsname}

\usepackage{subcaption}

\usepackage{algorithmic}

\title{Deep Sub-Ensembles for Fast Uncertainty Estimation in Image Classification}

\author{%
  Matias Valdenegro-Toro\\
  German Research Center for Artificial Intelligence\\
  28359 Bremen, Germany \\
  \texttt{matias.valdenegro@dfki.de} \\
}

\begin{document}

\maketitle

\begin{abstract}
    Fast estimates of model uncertainty are required for many robust robotics applications. Deep Ensembles provides state of the art uncertainty without requiring Bayesian methods, but still it is computationally expensive. In this paper we propose deep sub-ensembles, an approximation to deep ensembles where the core idea is to ensemble only the layers close to the output, and not the whole model. With ResNet-20 on the CIFAR10 dataset, we obtain 1.5-2.5 speedup over a Deep Ensemble, with a small increase in error and NLL, and similarly up to 5-15 speedup with a VGG-like network on the SVHN dataset. Our results show that this idea enables a trade-off between error and uncertainty quality versus computational performance.
\end{abstract}

\section{Introduction}

Neural networks have revolutionized many fields like object detection, behavior learning, and natural language processing. But most neural network models are overconfident, producing predictions that do not consider epistemic uncertainty, and are generally not calibrated. 

Many methods exist to augment neural networks with epistemic uncertainty, for example MC-Dropout \cite{gal2016dropout} and Deep Ensembles \cite{lakshminarayanan2017simple}. In particular the latter method is a good candidate for many applications due to simplicity and quality of uncertainty. For robotics applications, fast (close to real-time) estimates of uncertainty are highly desirable \cite{sunderhauf2018limits}.

In this paper we propose a simplification of the Deep Ensembles method. By only ensembling part of the model, while sharing a common network trunk, we show that an ensemble model still produces high quality uncertainty estimates in image classification tasks. This allows for a much faster inference time as a single pass is required for a large trunk network, and several forward passes for the sub-models connected to the output.

\section{Deep Sub-Ensembles}

Deep Ensembles \cite{lakshminarayanan2017simple} is  a non-Bayesian method for uncertainty quantification of machine learning models. It has been shown that an ensemble of models can produce good estimates of uncertainty, even surpassing methods like MC-Dropout.

Training and performing inference in a Deep Ensemble is computationally expensive. We consider that a neural network architecture can be logically divided \cite{sharif2014cnn} into two sub-networks, the trunk network $T$, and the task network $K$. The full architecture output for an input $x$ is then $K(T(x))$.

A Deep Sub-Ensemble conceptually corresponds to training one instance of the full network $K(T(x))$ in a training set, and then fixing the trunk network weights ($T_f$), and training additional instances of the model where only the task weights are learned. An overview of the training and inference process is shown in Figure \ref{trainingAlgorithm}. Our concept of a Deep Sub-Ensemble is similar to Bootstrapped DQN \cite{osband2016deep}, where a shared network and multiple output heads are used to produce high quality uncertainty estimates, but in a Deep Sub-Ensemble no bootstrap estimates are used, each combination of fixed trunk and trainable task network is trained on the same full dataset.

The purpose of this method is to allow the construction of an ensemble that contains a common trunk network $T_f$, and several instances of the task network $K_i$, making the ensemble computationally less expensive to evaluate at inference time, as generally the trunk network contains more computation than the task networks, and the trunk network is evaluated once. For classification, the sub-ensemble output is the average of task network probability predictions, namely $f(x) = N^{-1} \sum_i K_i(T_f(x))$.

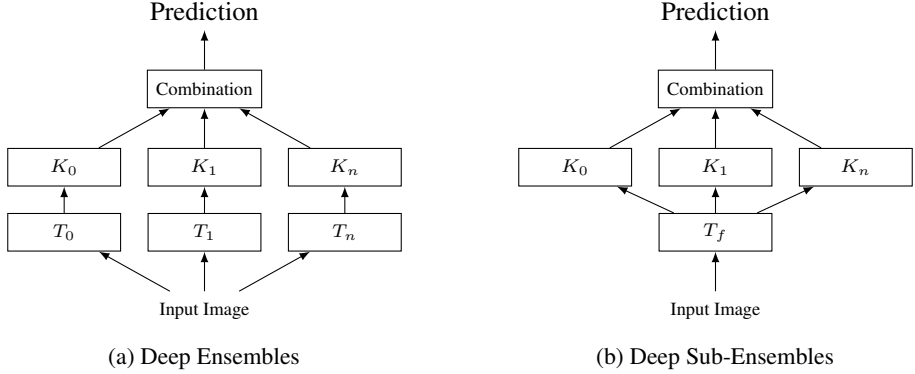
\begin{figure}
    \centering
    \begin{subfigure}{0.48\textwidth}
        \centering
        \begin{tikzpicture}[style={align=center, minimum height=0.5cm, minimum width = 1.5cm}]
        \node[] (dummy) {Prediction};
        
        \node[draw, below=1.5em of dummy] (comb) {{\scriptsize Combination}};
        
        \node[draw, below=1.5em of comb](k1) {\scriptsize{$K_1$}};
        \node[draw, below=1em of k1](t1) {\scriptsize{$T_1$}};
        \node[below=1.5em of t1](inputImage) {{\scriptsize Input Image}};
        \draw[-latex] (inputImage) -- (t1);
        \draw[-latex] (t1) -- (k1);
        \draw[-latex] (k1) -- (comb);
        \draw[-latex] (comb) -- (dummy);
        
        \node[draw, left=1em of k1](k0) {\scriptsize{$K_0$}};
        \node[draw, below=1em of k0](t0) {\scriptsize{$T_0$}};
        \draw[-latex] (inputImage) -- (t0);
        \draw[-latex] (t0) -- (k0);
        \draw[-latex] (k0) -- (comb);
        
        \node[draw, right=1em of k1](kn) {\scriptsize{$K_n$}};
        \node[draw, below=1em of kn](tn) {\scriptsize{$T_n$}};
        \draw[-latex] (inputImage) -- (tn);
        \draw[-latex] (tn) -- (kn);
        \draw[-latex] (kn) -- (comb);   
        \end{tikzpicture}  
        \caption{Deep Ensembles}
    \end{subfigure}
    \begin{subfigure}{0.48\textwidth}
        \centering
        \begin{tikzpicture}[style={align=center, minimum height=0.5cm, minimum width = 1.5cm}]
        \node[] (dummy) {Prediction};
        
        \node[draw, below=1.5em of dummy] (comb) {{\scriptsize Combination}};
        
        \node[draw, below=1.5em of comb](k1) {\scriptsize{$K_1$}};
        \node[draw, below=1em of k1](t1) {\scriptsize{$T_f$}};
        \node[below=1.5em of t1](inputImage) {{\scriptsize Input Image}};
        \draw[-latex] (inputImage) -- (t1);
        \draw[-latex] (t1) -- (k1);
        \draw[-latex] (k1) -- (comb);
        \draw[-latex] (comb) -- (dummy);
        
        \node[draw, left=1em of k1](k0) {\scriptsize{$K_0$}};
        \draw[-latex] (t1) -- (k0);
        \draw[-latex] (k0) -- (comb);
        
        \node[draw, right=1em of k1](kn) {\scriptsize{$K_n$}};
        \draw[-latex] (t1) -- (kn);
        \draw[-latex] (kn) -- (comb);   
        \end{tikzpicture}  
        \caption{Deep Sub-Ensembles}
    \end{subfigure}
    \caption{Conceptual comparison of Deep Ensembles and Deep Sub-Ensembles with $n$ ensemble members. The figure shows that in the latter, only a single trunk network $T_f$ is shared across all ensemble members, while in the former multiple trunk networks $T_i$ are used. In both cases the ensemble predictions are combined to produce outputs with uncertainty.}    
\end{figure}

\begin{figure}
    \begin{algorithmic}[1]
        \REQUIRE Training set $D$, Trunk and Task models $T$ and $K$, number of ensemble members $n$.
        \ENSURE Trained trunk network $T_f$ and ensemble members $E$.
        \STATE Stack the trunk and task model $T$-$K$ and train an initial instance of it on $D$.
        \STATE Freeze weights of the initially trained instance of $T$, producing $T_f$
        \STATE Set ensemble $E = \{ K \}$ with the initially trained instance of $K$
        \FOR{$i=1$ to $i=n-1$}
            \STATE Stack $T_f$ and a randomly initialized instance of $K$ and train it on $D$.
            \STATE $E = E \cup K$
        \ENDFOR
        \STATE Ensemble predictions can now be made by evaluating $T_f$ with an input image, then evaluating each ensemble member in $E$ given the output of $T_f$, and combining the predictions.
    \end{algorithmic}
    \caption{Training and Inference process for Deep Sub-Ensembles}
    \label{trainingAlgorithm}
\end{figure}

\section{Experimental Results in Image Classification}

We evaluate our proposed method in three datasets for image classification: MNIST, CIFAR10, and SVHN. For MNIST \cite{lecun1998mnist}, we use a simple batch normalized CNN consisting of a 32 $3 \times 3$ convolution, followed by 64 $3 \times 3$ convolution, and a fully connected layer with 128 neurons and an output fully connected layer with 10 neurons and a softmax activation. All layers use ReLU activations. We select two sets of task networks for ensembling, the first uses the last two fully connected layers (denominated SE-1), and the second task network uses the three last layers (2 FC and one Conv, denominated SE-2). These results are shown in Figure \ref{mnistResults}.

For CIFAR10 \cite{krizhevsky2009learning}, we use ResNet-20 \cite{he2016deep} with random shifts and horizontal flips as data augmentation. We define a set of two task networks, the first containing the classification layers and the last ResNet stack (with 64 filters and stride $S = 2$, denominated SE-1), and a second task network containing the previously defined network plus the second from last ResNet stack (with 32 filters and $S = 2$, denominated SE-2). These results are shown in Figure \ref{cifar10Results}.

Finally, for SVHN \cite{netzer2011reading} we use a batch normalized VGG-like network \cite{simonyan2014very}, with modules defined as two convolutional layers with the same number of filters and ReLU activation, and one $2 \times 2$ max pooling layer. The network is composed of modules with 32, 64, 128, and 128 filters, and followed by a fully connected layer of 128 neurons, and a final output layer with 10 neurons and softmax activation. We define a set of four task networks that we evaluate, namely taking the classification layers, and going from these layers backwards through the network modules, denominated as SE-1 to SE-4. These results are shown in Figure \ref{svhnResults}.

\subsection{Error and Uncertainty Quality}

For all datasets we evaluate both the classification error, and the negative log-likelihood. For SVHN we additionally evaluate the calibration curve. We compare our proposed method (called Deep Sub-Ensembles, SE) with Deep Ensembles \cite{lakshminarayanan2017simple} (DE), as the number of ensemble members is varied, from 1 to 15 ensemble members and task networks.

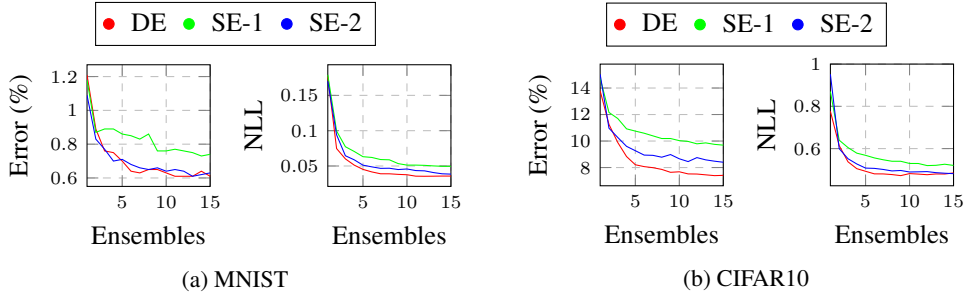
\begin{figure}
    \centering
    \begin{subfigure}{0.48\textwidth}    
       \centering
       \begin{tikzpicture}
        \begin{customlegend}[legend columns = 3,legend style = {column sep=1ex}, legend cell align = left,
        legend entries={DE, SE-1, SE-2}]
        \addlegendimage{mark=none,red, only marks}
        \addlegendimage{mark=none,green, only marks}
        \addlegendimage{mark=none,blue, only marks}
        \end{customlegend}
        \end{tikzpicture}        
        
        \vspace*{0.5em}
        
        \begin{tikzpicture}
        \begin{axis}[height = 0.14 \textheight, width = 0.48 \textwidth, xlabel={Ensembles}, ylabel={Error (\%)}, xmin = 1.0, xmax = 15.0, ymajorgrids=true, xmajorgrids=true, grid style=dashed, legend pos = north east, legend style={font=\scriptsize}, tick label style={font=\scriptsize}]
        
        \addplot+[mark = none, red] table[x  = num_ensembles, y  = error, col sep = semicolon] {experiments/mnist/deepensembles_cnn_mnist.csv};
        
        \addplot+[mark = none, green] table[x  = num_ensembles, y  = error, col sep = semicolon] {experiments/mnist/sub-deepensembles_1_cnn_mnist.csv};
        
        \addplot+[mark = none, blue] table[x  = num_ensembles, y  = error, col sep = semicolon] {experiments/mnist/sub-deepensembles_2_cnn_mnist.csv};        
        \end{axis}		
        \end{tikzpicture}    
        \begin{tikzpicture}
        \begin{axis}[height = 0.14 \textheight, width = 0.48 \textwidth, xlabel={Ensembles}, ylabel={NLL}, xmin = 1.0, xmax = 15.0, ymajorgrids=true, xmajorgrids=true, yticklabel style={/pgf/number format/fixed}, grid style=dashed, legend pos = north east, legend style={font=\scriptsize}, tick label style={font=\scriptsize}]
        
        \addplot+[mark = none, red] table[x  = num_ensembles, y  = nll, col sep = semicolon] {experiments/mnist/deepensembles_cnn_mnist.csv};
        
        \addplot+[mark = none, green] table[x  = num_ensembles, y  = nll, col sep = semicolon] {experiments/mnist/sub-deepensembles_1_cnn_mnist.csv};
        
        \addplot+[mark = none, blue] table[x  = num_ensembles, y  = nll, col sep = semicolon] {experiments/mnist/sub-deepensembles_2_cnn_mnist.csv};
        
        \end{axis}		
        \end{tikzpicture}        
        \caption{MNIST}
        \label{mnistResults}
    \end{subfigure}
    \begin{subfigure}{0.48\textwidth}
        \centering
        \begin{tikzpicture}columns
        \begin{customlegend}[legend columns = 3,legend style = {column sep=1ex}, legend cell align = left,
        legend entries={DE, SE-1, SE-2}]
        \addlegendimage{mark=none,red, only marks}
        \addlegendimage{mark=none,green, only marks}
        \addlegendimage{mark=none,blue, only marks}
        \end{customlegend}
        \end{tikzpicture}
        \begin{tikzpicture}
        \begin{axis}[height = 0.14 \textheight, width = 0.48 \textwidth, xlabel={Ensembles}, ylabel={Error (\%)}, xmin = 1.0, xmax = 15.0, ymajorgrids=true, xmajorgrids=true, grid style=dashed, legend pos = north east, legend style={font=\scriptsize}, tick label style={font=\scriptsize}]
        
        \addplot+[mark = none, red] table[x  = num_ensembles, y  = error, col sep = semicolon] {experiments/cifar10/deepensembles_resnet_cifar10.csv};
        
        \addplot+[mark = none, green] table[x  = num_ensembles, y  = error, col sep = semicolon] {experiments/cifar10/sub-deepensembles_1_resnet_cifar10.csv};
        
        \addplot+[mark = none, blue] table[x  = num_ensembles, y  = error, col sep = semicolon] {experiments/cifar10/sub-deepensembles_2_resnet_cifar10.csv};
        
        \end{axis}		
        \end{tikzpicture}
        \begin{tikzpicture}
        \begin{axis}[height = 0.14 \textheight, width = 0.48 \textwidth, xlabel={Ensembles}, ylabel={NLL}, xmin = 1.0, xmax = 15.0, ymajorgrids=true, xmajorgrids=true, yticklabel style={/pgf/number format/fixed}, grid style=dashed, legend pos = north east, legend style={font=\scriptsize}, tick label style={font=\scriptsize}]
        
        \addplot+[mark = none, red] table[x  = num_ensembles, y  = nll, col sep = semicolon] {experiments/cifar10/deepensembles_resnet_cifar10.csv};
        
        \addplot+[mark = none, green] table[x  = num_ensembles, y  = nll, col sep = semicolon] {experiments/cifar10/sub-deepensembles_1_resnet_cifar10.csv};
        
        \addplot+[mark = none, blue] table[x  = num_ensembles, y  = nll, col sep = semicolon] {experiments/cifar10/sub-deepensembles_2_resnet_cifar10.csv};
        
        \end{axis}		
        \end{tikzpicture}
        \caption{CIFAR10}
        \label{cifar10Results}
    \end{subfigure}
    \caption{Results on MNIST (with a simple CNN) and CIFAR10 (with ResNet-20), showing error and negative log-likelihood as the number of ensembles is varied}
\end{figure}

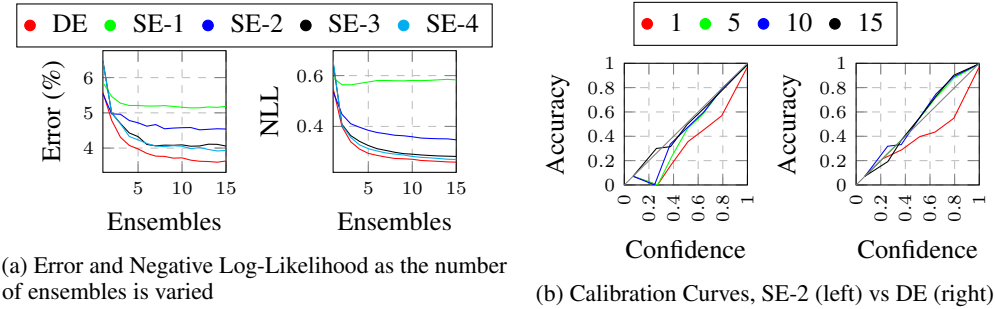
\begin{figure}
    \centering
    \begin{subfigure}{0.48\textwidth}    
    \centering
    \begin{tikzpicture}
    \begin{customlegend}[legend columns = 5,legend style = {column sep=1ex}, legend cell align = left,
    legend entries={DE, SE-1, SE-2, SE-3, SE-4}]
    \addlegendimage{mark=none,red, only marks}
    \addlegendimage{mark=none,green, only marks}
    \addlegendimage{mark=none,blue, only marks}
    \addlegendimage{mark=none,black, only marks}
    \addlegendimage{mark=none,cyan, only marks}
    \end{customlegend}
    \end{tikzpicture}    
    \begin{tikzpicture}
    \begin{axis}[height = 0.14 \textheight, width = 0.48 \textwidth, xlabel={Ensembles}, ylabel={Error (\%)}, xmin = 1.0, xmax = 15.0, ymajorgrids=true, xmajorgrids=true, grid style=dashed, legend pos = north east, legend style={font=\scriptsize}, tick label style={font=\scriptsize}]
    
    \addplot+[mark = none, red] table[x  = num_ensembles, y  = error, col sep = semicolon] {experiments/svhn/deepensembles_cnn_svhn.csv};
    
    \addplot+[mark = none, green] table[x  = num_ensembles, y  = error, col sep = semicolon] {experiments/svhn/sub-deepensembles_1_cnn_svhn.csv};
    
    \addplot+[mark = none, blue] table[x  = num_ensembles, y  = error, col sep = semicolon] {experiments/svhn/sub-deepensembles_2_cnn_svhn.csv};
    
    \addplot+[mark = none, black] table[x  = num_ensembles, y  = error, col sep = semicolon] {experiments/svhn/sub-deepensembles_3_cnn_svhn.csv};
    
    \addplot+[mark = none, cyan] table[x  = num_ensembles, y  = error, col sep = semicolon] {experiments/svhn/sub-deepensembles_4_cnn_svhn.csv};
    
    \end{axis}		
    \end{tikzpicture}    
    \begin{tikzpicture}
    \begin{axis}[height = 0.14 \textheight, width = 0.48\textwidth, xlabel={Ensembles}, ylabel={NLL}, xmin = 1.0, xmax = 15.0, ymajorgrids=true, xmajorgrids=true, yticklabel style={/pgf/number format/fixed}, grid style=dashed, legend pos = north east, legend style={font=\scriptsize}, tick label style={font=\scriptsize}]
    
    \addplot+[mark = none, red] table[x  = num_ensembles, y  = nll, col sep = semicolon] {experiments/svhn/deepensembles_cnn_svhn.csv};
    
    \addplot+[mark = none, green] table[x  = num_ensembles, y  = nll, col sep = semicolon] {experiments/svhn/sub-deepensembles_1_cnn_svhn.csv};
    
    \addplot+[mark = none, blue] table[x  = num_ensembles, y  = nll, col sep = semicolon] {experiments/svhn/sub-deepensembles_2_cnn_svhn.csv};
    
    \addplot+[mark = none, black] table[x  = num_ensembles, y  = nll, col sep = semicolon] {experiments/svhn/sub-deepensembles_3_cnn_svhn.csv};
    
    \addplot+[mark = none, cyan] table[x  = num_ensembles, y  = nll, col sep = semicolon] {experiments/svhn/sub-deepensembles_4_cnn_svhn.csv};
    
    \end{axis}		
    \end{tikzpicture}
    \caption{Error and Negative Log-Likelihood as the number of ensembles is varied}
    \label{svhnErrorNLL}
    \end{subfigure}
    \begin{subfigure}{0.48\textwidth}
        \centering
            \begin{tikzpicture}
            \begin{customlegend}[legend columns = 4,legend style = {column sep=1ex}, legend cell align = left,
            legend entries={1, 5, 10, 15}]
            \addlegendimage{mark=none,red, only marks}
            \addlegendimage{mark=none,green, only marks}
            \addlegendimage{mark=none,blue, only marks}
            \addlegendimage{mark=none,black, only marks}
            \end{customlegend}
            \end{tikzpicture}
            
            \begin{tikzpicture}
            \begin{axis}[height = 0.14 \textheight, width = 0.48 \textwidth, xlabel={Confidence}, ylabel={Accuracy}, xmin = 0.0, xmax = 1.0, ymin = 0.0, ymax = 1.0, ymajorgrids=true, xmajorgrids=true, grid style=dashed, legend pos = north east, legend style={font=\scriptsize}, x tick label style={font=\scriptsize, rotate=90}, y tick label style={font=\scriptsize}]
            
            \addplot+[mark = none, red] table[x  = conf, y  = acc, col sep = semicolon] {experiments/svhn/svhn-calibration-sub-deepensembles_2_num-ens-1_cnn_svhn.csv};
            
            \addplot+[mark = none, green] table[x  = conf, y  = acc, col sep = semicolon] {experiments/svhn/svhn-calibration-sub-deepensembles_2_num-ens-5_cnn_svhn.csv};
            
            \addplot+[mark = none, blue] table[x  = conf, y  = acc, col sep = semicolon] {experiments/svhn/svhn-calibration-sub-deepensembles_2_num-ens-10_cnn_svhn.csv};
            
            \addplot+[mark = none, black] table[x  = conf, y  = acc, col sep = semicolon] {experiments/svhn/svhn-calibration-sub-deepensembles_2_num-ens-15_cnn_svhn.csv};
            
            \addplot+[mark = none, gray] coordinates { (0, 0) (1, 1)};
            
            \end{axis}		
            \end{tikzpicture}
            \begin{tikzpicture}
            \begin{axis}[height = 0.14 \textheight, width = 0.48 \textwidth, xlabel={Confidence}, ylabel={Accuracy}, xmin = 0.0, xmax = 1.0, ymin = 0.0, ymax = 1.0, ymajorgrids=true, xmajorgrids=true, grid style=dashed, legend pos = north east, legend style={font=\scriptsize}, x tick label style={font=\scriptsize, rotate=90}, y tick label style={font=\scriptsize}]
            
            \addplot+[mark = none, red] table[x  = conf, y  = acc, col sep = semicolon] {experiments/svhn/svhn-calibration-deepensembles-num-ens-1_cnn_svhn.csv};
            
            \addplot+[mark = none, green] table[x  = conf, y  = acc, col sep = semicolon] {experiments/svhn/svhn-calibration-deepensembles-num-ens-5_cnn_svhn.csv};
            
            \addplot+[mark = none, blue] table[x  = conf, y  = acc, col sep = semicolon] {experiments/svhn/svhn-calibration-deepensembles-num-ens-10_cnn_svhn.csv};
            
            \addplot+[mark = none, black] table[x  = conf, y  = acc, col sep = semicolon] {experiments/svhn/svhn-calibration-deepensembles-num-ens-15_cnn_svhn.csv};
            
            \addplot+[mark = none, gray] coordinates { (0, 0) (1, 1)};
            
            \end{axis}		
            \end{tikzpicture}
            \caption{Calibration Curves, SE-2 (left) vs DE (right)}
            \label{svhnCalibrationCurves}
        \end{subfigure}
        \caption{Results on SVHN using a batch normalized VGG-like network}
        \label{svhnResults}
\end{figure}

On MNIST as shown in Figure \ref{mnistResults}, ensembling two layers of the model (SE-2) has error comparable with Deep Ensembles, but only ensembling the fully connected layers (SE-1) produces a higher error. The uncertainty as measured by the negative log-likelihood is comparable in all three scenarios, indicating the preliminar viability of our idea.

On CIFAR-10 (Figure \ref{cifar10Results}), error increases by around $2\%$ with a sub-ensemble when compared to Deep Ensembles, but the increase of NLL is minor, specially when ensembling two sets of layers (SE-2). Finally on SVHN (Figure \ref{svhnErrorNLL} ), there is a more marked difference in increasing error as less layers are ensembled. From SE-2 there is a clear improvement on negative log-likelihood, being very similar to the Deep Ensembles baseline since SE-3.

Calibration curves available in Figure \ref{svhnCalibrationCurves} show that both methods are calibrated, starting from being underconfident with the base model (single ensemble member), and with increasing confidence as ensemble members are added.

Overall our results show that Deep Sub-Ensembles is in all cases an approximation to Deep Ensembles, with always having higher error, but negative log-likelihood can be similar, depending on how many layers are ensembled. This is expected as ensembling less layers than the full model should behave as an approximation to the true ensemble, enabling a trade-off between computational resources and error and uncertainty quality.

\subsection{Trunk vs Ensemble Network Performance}

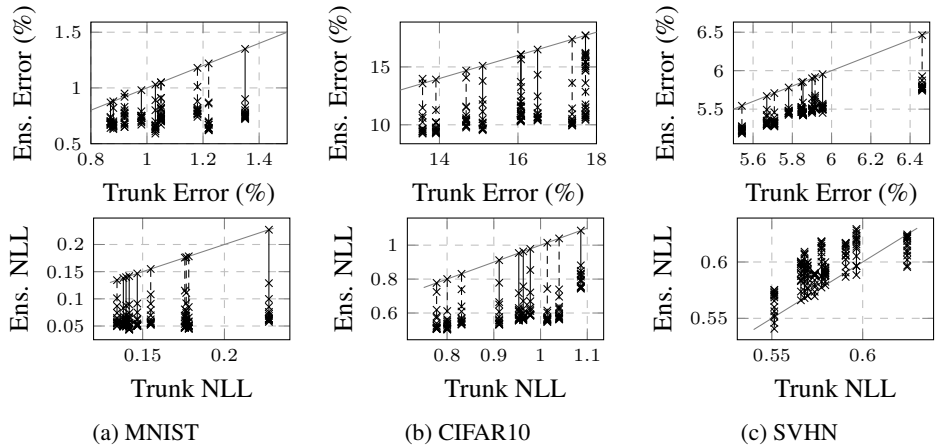
\begin{figure}
    \centering
    \begin{subfigure}{0.30 \textwidth}
        \centering
    \begin{tikzpicture}
        \begin{axis}[height = 0.14 \textheight, width = \textwidth, xlabel={Trunk Error (\%)}, ylabel={Ens. Error (\%)}, xmin = 0.8, xmax = 1.5, ymajorgrids=true, xmajorgrids=true, grid style=dashed, legend pos = north east, legend style={font=\scriptsize}, tick label style={font=\scriptsize}]
            
            \addplot+[mark = x, black] table[x  = base_error, y  = error, col sep = semicolon] {experiments/mnist/sub-deepensembles_1_cnn_mnist-run0.csv};
            
            \addplot+[mark = x, black] table[x  = base_error, y  = error, col sep = semicolon] {experiments/mnist/sub-deepensembles_1_cnn_mnist-run1.csv};
            
            \addplot+[mark = x, black] table[x  = base_error, y  = error, col sep = semicolon] {experiments/mnist/sub-deepensembles_1_cnn_mnist-run2.csv};
            
            \addplot+[mark = x, black] table[x  = base_error, y  = error, col sep = semicolon] {experiments/mnist/sub-deepensembles_1_cnn_mnist-run3.csv};
            
            \addplot+[mark = x, black] table[x  = base_error, y  = error, col sep = semicolon] {experiments/mnist/sub-deepensembles_1_cnn_mnist-run4.csv};
            
            \addplot+[mark = x, black] table[x  = base_error, y  = error, col sep = semicolon] {experiments/mnist/sub-deepensembles_1_cnn_mnist-run5.csv};
            
            \addplot+[mark = x, black] table[x  = base_error, y  = error, col sep = semicolon] {experiments/mnist/sub-deepensembles_1_cnn_mnist-run6.csv};
            
            \addplot+[mark = x, black] table[x  = base_error, y  = error, col sep = semicolon] {experiments/mnist/sub-deepensembles_1_cnn_mnist-run7.csv};
            
            \addplot+[mark = x, black] table[x  = base_error, y  = error, col sep = semicolon] {experiments/mnist/sub-deepensembles_1_cnn_mnist-run8.csv};
            
            \addplot+[mark = x, black] table[x  = base_error, y  = error, col sep = semicolon] {experiments/mnist/sub-deepensembles_1_cnn_mnist-run9.csv};
            
            \addplot+[mark = none, gray] coordinates { (0.8, 0.8) (1.5, 1.5)};                    
        \end{axis}		
    \end{tikzpicture}
    \begin{tikzpicture}
        \begin{axis}[height = 0.14 \textheight, width = \textwidth, xlabel={Trunk NLL}, ylabel={Ens. NLL}, ymajorgrids=true, xmajorgrids=true, yticklabel style={/pgf/number format/fixed}, grid style=dashed, legend pos = north east, legend style={font=\scriptsize}, tick label style={font=\scriptsize}]
        
        \addplot+[mark = x, black] table[x  = base_nll, y  = nll, col sep = semicolon] {experiments/mnist/sub-deepensembles_1_cnn_mnist-run0.csv};
        
        \addplot+[mark = x, black] table[x  = base_nll, y  = nll, col sep = semicolon] {experiments/mnist/sub-deepensembles_1_cnn_mnist-run1.csv};
        
        \addplot+[mark = x, black] table[x  = base_nll, y  = nll, col sep = semicolon] {experiments/mnist/sub-deepensembles_1_cnn_mnist-run2.csv};
        
        \addplot+[mark = x, black] table[x  = base_nll, y  = nll, col sep = semicolon] {experiments/mnist/sub-deepensembles_1_cnn_mnist-run3.csv};
        
        \addplot+[mark = x, black] table[x  = base_nll, y  = nll, col sep = semicolon] {experiments/mnist/sub-deepensembles_1_cnn_mnist-run4.csv};
        
        \addplot+[mark = x, black] table[x  = base_nll, y  = nll, col sep = semicolon] {experiments/mnist/sub-deepensembles_1_cnn_mnist-run5.csv};
        
        \addplot+[mark = x, black] table[x  = base_nll, y  = nll, col sep = semicolon] {experiments/mnist/sub-deepensembles_1_cnn_mnist-run6.csv};
        
        \addplot+[mark = x, black] table[x  = base_nll, y  = nll, col sep = semicolon] {experiments/mnist/sub-deepensembles_1_cnn_mnist-run7.csv};
        
        \addplot+[mark = x, black] table[x  = base_nll, y  = nll, col sep = semicolon] {experiments/mnist/sub-deepensembles_1_cnn_mnist-run8.csv};
        
        \addplot+[mark = x, black] table[x  = base_nll, y  = nll, col sep = semicolon] {experiments/mnist/sub-deepensembles_1_cnn_mnist-run9.csv};
        
        \addplot+[mark = none, gray] coordinates { (0.13, 0.13) (0.23, 0.23)};                    
        \end{axis}		
        \end{tikzpicture}
        \caption{MNIST}        
    \end{subfigure}
    \begin{subfigure}{0.30 \textwidth}
        \centering
        \begin{tikzpicture}
        \begin{axis}[height = 0.14 \textheight, width = \textwidth, xlabel={Trunk Error (\%)}, ylabel={Ens. Error (\%)}, xmin = 13.0, xmax = 18.0, ymajorgrids=true, xmajorgrids=true, grid style=dashed, legend pos = north east, legend style={font=\scriptsize}, tick label style={font=\scriptsize}]
        
        \addplot+[mark = x, black] table[x  = base_error, y  = error, col sep = semicolon] {experiments/cifar10/sub-deepensembles_1_resnet_cifar10-run0.csv};
        
        \addplot+[mark = x, black] table[x  = base_error, y  = error, col sep = semicolon] {experiments/cifar10/sub-deepensembles_1_resnet_cifar10-run1.csv};
        
        \addplot+[mark = x, black] table[x  = base_error, y  = error, col sep = semicolon] {experiments/cifar10/sub-deepensembles_1_resnet_cifar10-run2.csv};
        
        \addplot+[mark = x, black] table[x  = base_error, y  = error, col sep = semicolon] {experiments/cifar10/sub-deepensembles_1_resnet_cifar10-run3.csv};
        
        \addplot+[mark = x, black] table[x  = base_error, y  = error, col sep = semicolon] {experiments/cifar10/sub-deepensembles_1_resnet_cifar10-run4.csv};
        
        \addplot+[mark = x, black] table[x  = base_error, y  = error, col sep = semicolon] {experiments/cifar10/sub-deepensembles_1_resnet_cifar10-run5.csv};
        
        \addplot+[mark = x, black] table[x  = base_error, y  = error, col sep = semicolon] {experiments/cifar10/sub-deepensembles_1_resnet_cifar10-run6.csv};
        
        \addplot+[mark = x, black] table[x  = base_error, y  = error, col sep = semicolon] {experiments/cifar10/sub-deepensembles_1_resnet_cifar10-run7.csv};
        
        \addplot+[mark = x, black] table[x  = base_error, y  = error, col sep = semicolon] {experiments/cifar10/sub-deepensembles_1_resnet_cifar10-run8.csv};
        
        \addplot+[mark = x, black] table[x  = base_error, y  = error, col sep = semicolon] {experiments/cifar10/sub-deepensembles_1_resnet_cifar10-run9.csv};
        
        \addplot+[mark = none, gray] coordinates { (13.0, 13.0) (18.0, 18.0)};                    
        
        \end{axis}		
        \end{tikzpicture}
        \begin{tikzpicture}
        \begin{axis}[height = 0.14 \textheight, width = \textwidth, xlabel={Trunk NLL}, ylabel={Ens. NLL}, ymajorgrids=true, xmajorgrids=true, yticklabel style={/pgf/number format/fixed}, grid style=dashed, legend pos = north east, legend style={font=\scriptsize}, tick label style={font=\scriptsize}]

        \addplot+[mark = x, black] table[x  = base_nll, y  = nll, col sep = semicolon] {experiments/cifar10/sub-deepensembles_1_resnet_cifar10-run0.csv};
        
        \addplot+[mark = x, black] table[x  = base_nll, y  = nll, col sep = semicolon] {experiments/cifar10/sub-deepensembles_1_resnet_cifar10-run1.csv};
        
        \addplot+[mark = x, black] table[x  = base_nll, y  = nll, col sep = semicolon] {experiments/cifar10/sub-deepensembles_1_resnet_cifar10-run2.csv};
        
        \addplot+[mark = x, black] table[x  = base_nll, y  = nll, col sep = semicolon] {experiments/cifar10/sub-deepensembles_1_resnet_cifar10-run3.csv};
        
        \addplot+[mark = x, black] table[x  = base_nll, y  = nll, col sep = semicolon] {experiments/cifar10/sub-deepensembles_1_resnet_cifar10-run4.csv};
        
        \addplot+[mark = x, black] table[x  = base_nll, y  = nll, col sep = semicolon] {experiments/cifar10/sub-deepensembles_1_resnet_cifar10-run5.csv};
        
        \addplot+[mark = x, black] table[x  = base_nll, y  = nll, col sep = semicolon] {experiments/cifar10/sub-deepensembles_1_resnet_cifar10-run6.csv};
        
        \addplot+[mark = x, black] table[x  = base_nll, y  = nll, col sep = semicolon] {experiments/cifar10/sub-deepensembles_1_resnet_cifar10-run7.csv};
        
        \addplot+[mark = x, black] table[x  = base_nll, y  = nll, col sep = semicolon] {experiments/cifar10/sub-deepensembles_1_resnet_cifar10-run8.csv};
        
        \addplot+[mark = x, black] table[x  = base_nll, y  = nll, col sep = semicolon] {experiments/cifar10/sub-deepensembles_1_resnet_cifar10-run9.csv};
        
        \addplot+[mark = none, gray] coordinates { (0.75, 0.75) (1.1, 1.1)};
                
        \end{axis}		
        \end{tikzpicture}
        \caption{CIFAR10}        
    \end{subfigure}
    \begin{subfigure}{0.30 \textwidth}
        \centering
        \begin{tikzpicture}
        \begin{axis}[height = 0.14 \textheight, width = \textwidth, xlabel={Trunk Error (\%)}, ylabel={Ens. Error (\%)}, xmin = 5.5, xmax = 6.5, ymajorgrids=true, xmajorgrids=true, grid style=dashed, legend pos = north east, legend style={font=\scriptsize}, tick label style={font=\scriptsize}]
        
        \addplot+[mark = x, black] table[x  = base_error, y  = error, col sep = semicolon] {experiments/svhn/sub-deepensembles_1_cnn_svhn-run0.csv};
        
        \addplot+[mark = x, black] table[x  = base_error, y  = error, col sep = semicolon] {experiments/svhn/sub-deepensembles_1_cnn_svhn-run1.csv};
        
        \addplot+[mark = x, black] table[x  = base_error, y  = error, col sep = semicolon] {experiments/svhn/sub-deepensembles_1_cnn_svhn-run2.csv};
        
        \addplot+[mark = x, black] table[x  = base_error, y  = error, col sep = semicolon] {experiments/svhn/sub-deepensembles_1_cnn_svhn-run3.csv};
        
        \addplot+[mark = x, black] table[x  = base_error, y  = error, col sep = semicolon] {experiments/svhn/sub-deepensembles_1_cnn_svhn-run4.csv};
        
        \addplot+[mark = x, black] table[x  = base_error, y  = error, col sep = semicolon] {experiments/svhn/sub-deepensembles_1_cnn_svhn-run5.csv};
        
        \addplot+[mark = x, black] table[x  = base_error, y  = error, col sep = semicolon] {experiments/svhn/sub-deepensembles_1_cnn_svhn-run6.csv};
        
        \addplot+[mark = x, black] table[x  = base_error, y  = error, col sep = semicolon] {experiments/svhn/sub-deepensembles_1_cnn_svhn-run7.csv};
        
        \addplot+[mark = x, black] table[x  = base_error, y  = error, col sep = semicolon] {experiments/svhn/sub-deepensembles_1_cnn_svhn-run8.csv};
        
        \addplot+[mark = x, black] table[x  = base_error, y  = error, col sep = semicolon] {experiments/svhn/sub-deepensembles_1_cnn_svhn-run9.csv};
        
        \addplot+[mark = none, gray] coordinates { (5.5, 5.5) (6.5, 6.5)};
        
        \end{axis}		
        \end{tikzpicture}
        \begin{tikzpicture}
        \begin{axis}[height = 0.14 \textheight, width = \textwidth, xlabel={Trunk NLL}, ylabel={Ens. NLL}, ymajorgrids=true, xmajorgrids=true, yticklabel style={/pgf/number format/fixed}, grid style=dashed, legend pos = north east, legend style={font=\scriptsize}, tick label style={font=\scriptsize}]
        
        \addplot+[mark = x, black] table[x  = base_nll, y  = nll, col sep = semicolon] {experiments/svhn/sub-deepensembles_1_cnn_svhn-run0.csv};
        
        \addplot+[mark = x, black] table[x  = base_nll, y  = nll, col sep = semicolon] {experiments/svhn/sub-deepensembles_1_cnn_svhn-run1.csv};
        
        \addplot+[mark = x, black] table[x  = base_nll, y  = nll, col sep = semicolon] {experiments/svhn/sub-deepensembles_1_cnn_svhn-run2.csv};
        
        \addplot+[mark = x, black] table[x  = base_nll, y  = nll, col sep = semicolon] {experiments/svhn/sub-deepensembles_1_cnn_svhn-run3.csv};
        
        \addplot+[mark = x, black] table[x  = base_nll, y  = nll, col sep = semicolon] {experiments/svhn/sub-deepensembles_1_cnn_svhn-run4.csv};
        
        \addplot+[mark = x, black] table[x  = base_nll, y  = nll, col sep = semicolon] {experiments/svhn/sub-deepensembles_1_cnn_svhn-run5.csv};
        
        \addplot+[mark = x, black] table[x  = base_nll, y  = nll, col sep = semicolon] {experiments/svhn/sub-deepensembles_1_cnn_svhn-run6.csv};
        
        \addplot+[mark = x, black] table[x  = base_nll, y  = nll, col sep = semicolon] {experiments/svhn/sub-deepensembles_1_cnn_svhn-run7.csv};
        
        \addplot+[mark = x, black] table[x  = base_nll, y  = nll, col sep = semicolon] {experiments/svhn/sub-deepensembles_1_cnn_svhn-run8.csv};
        
        \addplot+[mark = x, black] table[x  = base_nll, y  = nll, col sep = semicolon] {experiments/svhn/sub-deepensembles_1_cnn_svhn-run9.csv};
        
        \addplot+[mark = none, gray] coordinates { (0.54, 0.54) (0.63, 0.63)};
        
        \end{axis}		
        \end{tikzpicture}
        \caption{SVHN}        
    \end{subfigure}
    \caption{Relationship between Sub-Ensemble and trunk network performance, in terms of error and negative log-likelihood. Here we only evaluate SE-1.}
    \label{trunkErrorNLLCorrelation}
\end{figure}

One additional property of a Deep Sub-Ensemble is that since a trunk network is trained once, due to random weight initialization and randomness in the training process, the model might not produce the best features given the data. We evaluated this by training 10 runs of Deep Sub-Ensembles, and evaluating the trunk model error and the ensemble error, these results are shown in Figure \ref{trunkErrorNLLCorrelation}. It can be seen that there is a strong correlation between trunk and ensemble error, with the same effect happening for negative log-likelihood. This indicates that a more thorough design and training of the trunk model might be necessary for good ensemble performance.

\subsection{Out of Distribution Detection - SVHN vs CIFAR10}

We have also evaluated the out of distribution detection (ODD) capabilities of Deep Sub-Ensembles. For this we used the ensemble model trained on SVHN, and evaluated on the CIFAR10 test set for ODD examples, and in the SVHN test set for in-distribution (ID) examples, as the image sizes are compatible (both are $32 \times 32$), and there are no classes in common.

To decide if an example is out of distribution, we use the entropy of the ensemble probabilities:

$$H(x) = -\sum_{c \in C} f(x)_c \log f(x)_c$$

Then we put a threshold in the entropy to decide if an example is in-distribution or out-of-distribution. The idea is that in-distribution examples will have a low entropy, as certain class probabilities dominate the prediction, while out-of-distribution examples will have a uniform class probability distribution, which increases entropy.

\begin{figure}[!h]
    \centering
    \begin{tikzpicture}
    \begin{customlegend}[legend columns = 4,legend style = {column sep=1ex}, legend cell align = left,
    legend entries={1, 5, 10, 15}]
    \addlegendimage{mark=none,red, only marks}
    \addlegendimage{mark=none,green, only marks}
    \addlegendimage{mark=none,blue, only marks}hyo 
    \addlegendimage{mark=none,black, only marks}
    \end{customlegend}
    \end{tikzpicture}
    
    \begin{subfigure}{0.48\textwidth}
        \centering
        \begin{tikzpicture}
        \begin{axis}[height = 0.16 \textheight, width = 0.70 \textwidth, xlabel={False Positive Rate}, ylabel={True Positive Rate}, xmin = 0.0, xmax = 1.0, ymin = 0.0, ymax = 1.0, ymajorgrids=true, xmajorgrids=true, yticklabel style={/pgf/number format/fixed}, grid style=dashed, legend pos = north east, legend style={font=\scriptsize}, tick label style={font=\scriptsize}]
        
        \addplot+[mark = none, red] table[x  = fpr, y  = tpr, col sep = semicolon] {experiments/svhn/ood-roc-sub-deepensembles_1_num-ens-1_cnn_svhn.csv};
        
        %\addplot+[mark = none] table[x  = fpr, y  = tpr, col sep = semicolon] {experiments/svhn/ood-roc-sub-deepensembles_1_num-ens-2_cnn_svhn.csv};
        
        \addplot+[mark = none, green] table[x  = fpr, y  = tpr, col sep = semicolon] {experiments/svhn/ood-roc-sub-deepensembles_1_num-ens-5_cnn_svhn.csv};
        
        \addplot+[mark = none, blue] table[x  = fpr, y  = tpr, col sep = semicolon] {experiments/svhn/ood-roc-sub-deepensembles_1_num-ens-10_cnn_svhn.csv};
        
        \addplot+[mark = none, black] table[x  = fpr, y  = tpr, col sep = semicolon] {experiments/svhn/ood-roc-sub-deepensembles_1_num-ens-15_cnn_svhn.csv};
        
        %\addplot+[mark = none] table[x  = fpr, y  = tpr, col sep = semicolon] {experiments/svhn/ood-roc-sub-deepensembles_1_num-ens-15_cnn_svhn.csv};
        
        \addplot+[mark = none, gray] coordinates { (0.0, 0.0) (1.0, 1.0)};
        
        \end{axis}		
        \end{tikzpicture}
        \caption{Deep Sub-Ensembles}
    \end{subfigure}
    \begin{subfigure}{0.48\textwidth}
        \center
        \begin{tikzpicture}
        \begin{axis}[height = 0.16 \textheight, width = 0.70 \textwidth, xlabel={False Positive Rate}, ylabel={True Positive Rate}, xmin = 0.0, xmax = 1.0, ymin = 0.0, ymax = 1.0, ymajorgrids=true, xmajorgrids=true, yticklabel style={/pgf/number format/fixed}, grid style=dashed, legend pos = north east, legend style={font=\scriptsize}, tick label style={font=\scriptsize}]
        
        \addplot+[mark = none, red] table[x  = fpr, y  = tpr, col sep = semicolon] {experiments/svhn/ood-roc-deepensembles-num-ens-1_cnn_svhn.csv};
        
        %\addplot+[mark = none] table[x  = fpr, y  = tpr, col sep = semicolon] {experiments/svhn/ood-roc-sub-deepensembles_1_num-ens-2_cnn_svhn.csv};
        
        \addplot+[mark = none, green] table[x  = fpr, y  = tpr, col sep = semicolon] {experiments/svhn/ood-roc-deepensembles-num-ens-5_cnn_svhn.csv};
        
        \addplot+[mark = none, blue] table[x  = fpr, y  = tpr, col sep = semicolon] {experiments/svhn/ood-roc-deepensembles-num-ens-10_cnn_svhn.csv};
        
        \addplot+[mark = none, black] table[x  = fpr, y  = tpr, col sep = semicolon] {experiments/svhn/ood-roc-deepensembles-num-ens-15_cnn_svhn.csv};
        
        %\addplot+[mark = none] table[x  = fpr, y  = tpr, col sep = semicolon] {experiments/svhn/ood-roc-sub-deepensembles_1_num-ens-15_cnn_svhn.csv};
        
        \addplot+[mark = none, gray] coordinates { (0.0, 0.0) (1.0, 1.0)};
        
        \end{axis}
        \end{tikzpicture}
        \caption{Deep Ensembles}
    \end{subfigure}
    \caption{ROC Curves for out-of-distribution detection using entropy on SVHN vs CIFAR10.}
    \label{oodROCCurves}
\end{figure}
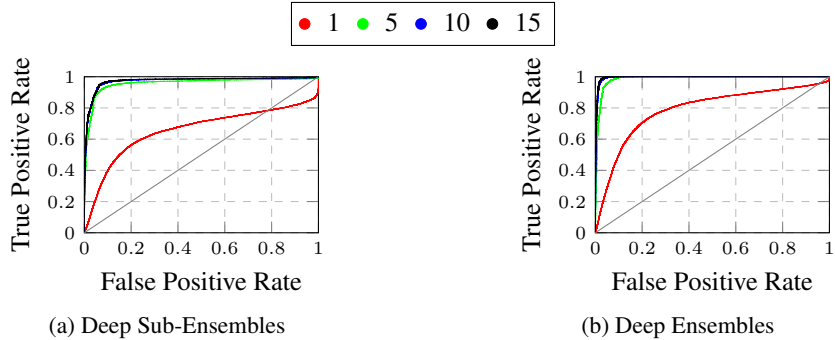

We evaluate performance of this method using the area under the ROC curve as the number of ensemble members is varied. Results are presented in Table \ref{oodNumericalResults} and Figure \ref{oodROCCurves}. Our results indicate that probabilities produced by Deep Ensembles have an excellent capability for out-of-distribution detection, starting from 5 ensemble members. Deep Sub-Ensembles also produces good separation between ID and OOD examples, but requires more ensemble members to reach performance that is slightly worse than a Deep Ensemble, at 15 ensemble members. The mean ID and OOD entropy show that it clearly divides ID and OOD examples.

\begin{table}[h]
    \centering
    \begin{tabular}{lllllll}
        \toprule
        & \multicolumn{3}{c}{Sub-Ensembles} & \multicolumn{3}{c}{Deep Ensembles}\\
        \midrule
        \# of Ensembles & AUC 	& ID Entropy 	& OOD Entropy & AUC 	& ID Entropy 	& OOD Entropy\\
        \midrule
        1				& 0.655	& 0.051				& 0.195			   & 0.787  & 0.054				& 0.281\\
        5				& 0.957	& 0.043				& 0.867			   & 0.986  & 0.112			    & 1.325\\
        10				& 0.971	& 0.042				& 0.986			   & 0.994  & 0.125				& 1.589\\
        15				& 0.973	& 0.040				& 1.009			   & 0.996  & 0.130				& 1.665\\
        \bottomrule
    \end{tabular}    
    \vspace*{0.5em}
    \caption{Numerical results for OOD detection on SVHN-CIFAR10. ID/OOD Entropy corresponds to sample means.}
    \label{oodNumericalResults}
\end{table}

\subsection{Computational Performance Analysis}

Ensembling less layers has a theoretical advantage over ensembling the full model. In this section we aim to evaluate this hypothesis and measure how much speedup can be obtained by using a sub-ensemble instead of a full deep ensemble.

For this purpose we estimate the number of floating point operations (FLOPs) \footnote{Not to be confused with FLOPS, which is floating point operations per second} for each architecture, as we vary the number of ensembles. We evaluate models for SVHN and CIFAR10, as they are the most complex ones. We plot the error and negative log-likelihood as function of FLOPS for different number of ensemble members, as a way to show the trade-off between error and uncertainty quality with computational requirements, and we also compute the speedup of a sub-ensemble over a full ensemble, computed as the number of FLOPs of a full ensemble divided by FLOPs for the sub-ensemble.

\begin{figure}[ht]
    \centering
    \begin{tikzpicture}
    \begin{customlegend}[legend columns = 3,legend style = {column sep=1ex}, legend cell align = left,
    legend entries={DE, SE-1, SE-2, SE-3, SE-4}]
    \addlegendimage{mark=none,red, only marks}
    \addlegendimage{mark=none,green, only marks}
    \addlegendimage{mark=none,blue, only marks}
    \end{customlegend}
    \end{tikzpicture}
    
    \begin{tikzpicture}
    \begin{axis}[height = 0.16 \textheight, width = 0.32 \textwidth, xlabel={FLOPs}, ylabel={Error (\%)}, ymajorgrids=true, xmajorgrids=true, grid style=dashed, legend pos = north east, legend style={font=\scriptsize}, tick label style={font=\scriptsize}]
    
    \addplot+[mark = none, red] table[x  = flops, y  = error, col sep = semicolon] {experiments/cifar10/deepensembles_resnet_cifar10.csv};
    
    \addplot+[mark = none, green] table[x  = flops, y  = error, col sep = semicolon] {experiments/cifar10/sub-deepensembles_1_resnet_cifar10.csv};
    
    \addplot+[mark = none, blue] table[x  = flops, y  = error, col sep = semicolon] {experiments/cifar10/sub-deepensembles_2_resnet_cifar10.csv};
        
    \end{axis}		
    \end{tikzpicture}    
    \begin{tikzpicture}
    \begin{axis}[height = 0.16 \textheight, width = 0.32\textwidth, xlabel={FLOPs}, ylabel={NLL}, ymajorgrids=true, xmajorgrids=true, yticklabel style={/pgf/number format/fixed}, grid style=dashed, legend pos = north east, legend style={font=\scriptsize}, tick label style={font=\scriptsize}]
    
    \addplot+[mark = none, red] table[x  = flops, y  = nll, col sep = semicolon] {experiments/cifar10/deepensembles_resnet_cifar10.csv};
    
    \addplot+[mark = none, green] table[x  = flops, y  = nll, col sep = semicolon] {experiments/cifar10/sub-deepensembles_1_resnet_cifar10.csv};
    
    \addplot+[mark = none, blue] table[x  = flops, y  = nll, col sep = semicolon] {experiments/cifar10/sub-deepensembles_2_resnet_cifar10.csv};
    
    \end{axis}		
    \end{tikzpicture}
    \begin{tikzpicture}
    \begin{axis}[height = 0.16 \textheight, width = 0.32\textwidth, xlabel={Ensembles}, ylabel={Speedup}, ymajorgrids=true, xmajorgrids=true, yticklabel style={/pgf/number format/fixed}, grid style=dashed, legend pos = north east, legend style={font=\scriptsize}, tick label style={font=\scriptsize}]
    
    \addplot+[mark = none, green] table[x  = num_ensembles, y  = speedup, col sep = semicolon] {experiments/cifar10/sub-deepensembles_cifar10_resnet_speedup-1.csv};
    
    \addplot+[mark = none, blue] table[x  = num_ensembles, y  = speedup, col sep = semicolon] {experiments/cifar10/sub-deepensembles_cifar10_resnet_speedup-2.csv};
    
    \end{axis}		
    \end{tikzpicture}
    \caption{Computational performance measured as FLOPs and Speedup on CIFAR10 using ResNet-20}
    \label{cifar10ComputationalComplexity}
\end{figure}
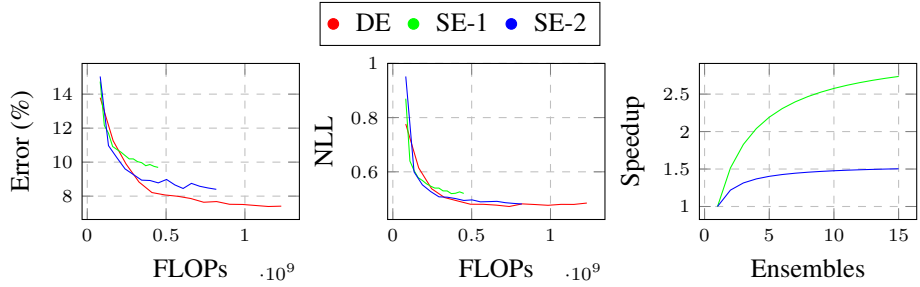

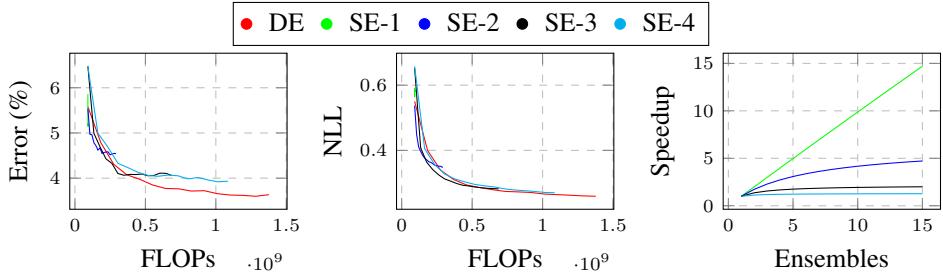
\begin{figure}[ht]
    \centering
        \begin{tikzpicture}
        \begin{customlegend}[legend columns = 5,legend style = {column sep=1ex}, legend cell align = left,
        legend entries={DE, SE-1, SE-2, SE-3, SE-4}]
        \addlegendimage{mark=none,red, only marks}
        \addlegendimage{mark=none,green, only marks}
        \addlegendimage{mark=none,blue, only marks}
        \addlegendimage{mark=none,black, only marks}
        \addlegendimage{mark=none,cyan, only marks}
        \end{customlegend}
        \end{tikzpicture}
         
        \begin{tikzpicture}
        \begin{axis}[height = 0.16 \textheight, width = 0.32 \textwidth, xlabel={FLOPs}, ylabel={Error (\%)}, ymajorgrids=true, xmajorgrids=true, grid style=dashed, legend pos = north east, legend style={font=\scriptsize}, tick label style={font=\scriptsize}]
        
        \addplot+[mark = none, red] table[x  = flops, y  = error, col sep = semicolon] {experiments/svhn/deepensembles_cnn_svhn.csv};
        
        \addplot+[mark = none, green] table[x  = flops, y  = error, col sep = semicolon] {experiments/svhn/sub-deepensembles_1_cnn_svhn.csv};
        
        \addplot+[mark = none, blue] table[x  = flops, y  = error, col sep = semicolon] {experiments/svhn/sub-deepensembles_2_cnn_svhn.csv};
        
        \addplot+[mark = none, black] table[x  = flops, y  = error, col sep = semicolon] {experiments/svhn/sub-deepensembles_3_cnn_svhn.csv};
        
        \addplot+[mark = none, cyan] table[x  = flops, y  = error, col sep = semicolon] {experiments/svhn/sub-deepensembles_4_cnn_svhn.csv};
        
        \end{axis}		
        \end{tikzpicture}    
        \begin{tikzpicture}
        \begin{axis}[height = 0.16 \textheight, width = 0.32\textwidth, xlabel={FLOPs}, ylabel={NLL}, ymajorgrids=true, xmajorgrids=true, yticklabel style={/pgf/number format/fixed}, grid style=dashed, legend pos = north east, legend style={font=\scriptsize}, tick label style={font=\scriptsize}]
        
        \addplot+[mark = none, red] table[x  = flops, y  = nll, col sep = semicolon] {experiments/svhn/deepensembles_cnn_svhn.csv};
        
        \addplot+[mark = none, green] table[x  = flops, y  = nll, col sep = semicolon] {experiments/svhn/sub-deepensembles_1_cnn_svhn.csv};
        
        \addplot+[mark = none, blue] table[x  = flops, y  = nll, col sep = semicolon] {experiments/svhn/sub-deepensembles_2_cnn_svhn.csv};
        
        \addplot+[mark = none, black] table[x  = flops, y  = nll, col sep = semicolon] {experiments/svhn/sub-deepensembles_3_cnn_svhn.csv};
        
        \addplot+[mark = none, cyan] table[x  = flops, y  = nll, col sep = semicolon] {experiments/svhn/sub-deepensembles_4_cnn_svhn.csv};
        
        \end{axis}		
        \end{tikzpicture}
        \begin{tikzpicture}
        \begin{axis}[height = 0.16 \textheight, width = 0.32\textwidth, xlabel={Ensembles}, ylabel={Speedup}, ymajorgrids=true, xmajorgrids=true, yticklabel style={/pgf/number format/fixed}, grid style=dashed, legend pos = north east, legend style={font=\scriptsize}, tick label style={font=\scriptsize}]
        
        \addplot+[mark = none, green] table[x  = num_ensembles, y  = speedup, col sep = semicolon] {experiments/svhn/sub-deepensembles_svhn_cnn_speedup-1.csv};
        
        \addplot+[mark = none, blue] table[x  = num_ensembles, y  = speedup, col sep = semicolon] {experiments/svhn/sub-deepensembles_svhn_cnn_speedup-2.csv};
        
        \addplot+[mark = none, black] table[x  = num_ensembles, y  = speedup, col sep = semicolon] {experiments/svhn/sub-deepensembles_svhn_cnn_speedup-3.csv};
        
        \addplot+[mark = none, cyan] table[x  = num_ensembles, y  = speedup, col sep = semicolon] {experiments/svhn/sub-deepensembles_svhn_cnn_speedup-4.csv};
        
        \end{axis}		
        \end{tikzpicture}
    \caption{Computational performance measured as FLOPs and Speedup on SVHN using a batch normalized VGG-like network}
    \label{svhnComputationalComplexity}
\end{figure}

Results are shown in Figure \ref{cifar10ComputationalComplexity} for CIFAR10 with ResNet-20, and \ref{svhnComputationalComplexity} for SVHN with a batch normalized VGG-like network. For CIFAR10, there is a clear trade-off between error and computation, but it is possible to trade small NLL amounts for big gains in computational performance (up to 1.5-2.5 times).

For SVHN, similar patterns in error trade-offs are seen, and the NLL decreases considerably with small compute, for example SE-1 almost has no gains in NLL with very small increases in FLOPs. SE-2 and SE-3 allow to trade small variations in NLL for large computational gains (up to 2-5 times).

Looking at speedups it can be seen that a sub-ensemble obtains decent speed improvements over a full ensemble, but there are large variations in speedup depending on model complexity and number of ensemble members. For example a maximum speedup of 15 can be reached with SE-1 on a VGG-like network, but smaller speedups are obtained on ResNet, maximum 2.7 with SE-2. It is clear that speedups heavily depend on the granularity and selection of layers to be ensembled.

\section{Conclusions and Future Work}

In this paper we have presented deep sub-ensembles for image classification, a simplification of deep ensembles with the purpose of reducing computation time at inference.

Our preliminary results show that it might not be necessary to ensemble all the layers in a model, and that a trade-off between computation time and uncertainty quality might be possible, depending on the task and dataset being learned. In terms of FLOPs, we measured speedups up to 1.5-2.5 for ResNet-20 on the CIFAR10 dataset, and speedups of 5-15 for a VGG-like network on the SVHN dataset, with small increase in error and NLL.

As future work, we wish to evaluate on the ImageNet dataset, and explore ways to train sub-ensembles in an end-to-end fashion, which could reduce training time. We will also extend this evaluation to regression and estimate computation times on CPU and GPU.

\section*{Acknowledgements}

This work has been partially supported by the Autonomous Harbour Cleaning project funded by EIT Digital (Ref 18181).

\newpage
\bibliographystyle{plain}
\bibliography{biblio}

\end{document}